\documentclass[lettersize,journal]{IEEEtran}
\IEEEoverridecommandlockouts
\usepackage{amsmath,amssymb,amsthm,url,mathtools}
\usepackage{graphicx,xcolor,subcaption}
\usepackage{algorithm,algpseudocode,latexsym,enumitem}%
\usepackage{physics}
  
\algnewcommand{\Inputs}[1]{%
  \State \textbf{Inputs:} #1
}
\algnewcommand{\Initialize}[1]{%
  \State \textbf{Initialize:}
  \Statex \hspace*{\algorithmicindent}\parbox[t]{.8\linewidth}{\raggedright #1}
}

\newtheorem{definition}{Definition}

\begin{document}

\title{Driver Fatigue Prediction using Randomly Activated Neural Networks for Smart Ridesharing Platforms}

\author{
\IEEEauthorblockN{Sree Pooja Akula, Mukund Telukunta, Venkata Sriram Siddhardh Nadendla} 
\\[1.5ex]
\IEEEauthorblockA{Department of Computer Science, \\ Missouri University of Science and Technology, Rolla, MO, USA.\\
Email: \{sacqx, mt3qb, nadendla\}@umsystem.edu}
}


\maketitle

\begin{abstract}
Drivers in ridesharing platforms exhibit cognitive atrophy and fatigue as they accept ride offers along the day, which can have a significant impact on the overall efficiency of the ridesharing platform. In contrast to the current literature which focuses primarily on modeling and learning driver's preferences across different ride offers, this paper proposes a novel \emph{\textbf{D}ynamic \textbf{D}iscounted \textbf{S}atisficing} (DDS) heuristic to model and predict driver's sequential ride decisions during a given shift. Based on DDS heuristic, a novel \emph{stochastic neural network with random activations} is proposed to model DDS heuristic and predict the final decision made by a given driver. The presence of random activations in the network necessitated the development of a novel training algorithm called \emph{Sampling-Based Back Propagation Through Time (SBPTT)}, where gradients are computed for independent instances of neural networks (obtained via sampling the distribution of activation threshold) and aggregated to update the network parameters. Using both simulation experiments as well as on real Chicago taxi dataset,
this paper demonstrates the improved performance of the proposed approach, when compared to state-of-the-art methods. 
\end{abstract}


\section{Introduction}

Ridesharing platform is a cyber-physical-human system where available taxi drivers are matched to ride-seeking passengers by the platform, based on spatio-temporal dynamics of diverse factors such as traffic congestion, ride availability and revenue. 
In traditional taxi services, drivers are expected to serve passengers who are assigned to them by the platform without any choice. On the other hand, modern ridesharing platforms (e.g. Uber and Lyft) match drivers with passengers and allow drivers to either accept/reject the match offer, thus preserving the decision autonomy at both types of agents \cite{aguilera2022ridesourcing,vega2023traditional}. 
However, if the ridesharing platform can predict the stopping decision made by every driver prior to that actual event, then ride requests can be optimized so as to maximize the average revenue of the platform. However, little work is available on the prediction of driver's stopping decision (i.e. final task performed during a given day), especially when they exhibit behavioral deviations from expected utility maximization (EUM) behavior \cite{green1986expected}. Therefore, this paper focuses on the prediction of driver's stopping decision to help improve the performance of ridesharing platforms.

Practical ridesharing platforms offer diverse recommendations to both passengers as well as drivers. 
For example, passengers are provided with wait-location recommendation to reduce the trip-cost \cite{dai2016ridesharing} and plan ride suggestions \cite{svangren2018passenger} to improve their experience. On the other hand, drivers are provided with ride choices, along with incentives if/when passenger's future activity is predicted in a location that has very few drivers. The success of such recommendations relies heavily on the accuracy of \emph{network state information} (NSI) \cite{shu2020spatial,zhang2021taxiint} available at the platform. For example, Altshuler \emph{et al.} predicted spatio-temporal utilization of ridesharing services from passenger activity models extracted from NSI \cite{altshuler2019modeling}. \cite{hu2018taxi} extracted NSI from GPS data and identified the passenger's demand hot area and proposed a taxi station optimization model by analyzing the time series distribution dynamic characteristics of passenger's temporal variation in certain land use types and taxi driver’s searching behavior in connection with different activity spaces for different lengths of observation period. 

However, all of the aforementioned works rely on NSI, and do not capture the attributes related to driver's behavior. Macadam in \cite{macadam2003} emphasizes the importance of including human characteristics in models of driver control behavior to accurately predict the performance of the driver-vehicle system. Macadam identified physical limitations and unique attributes of human drivers and presented driver models commonly used for prediction. 
A natural surrogate to driver's cognitive atrophy due to fatigue is their \emph{stopping task} (a.k.a. the total number of rides completed by the driver).
Therefore, predicting driver's stopping task can greatly help in 
improving the performance of ridesharing platforms.
However, to the best of our knowledge, there is little work on cognitive atrophy prediction from driver's ride productivity on a ridesharing platform. Devaguptapu in \cite{Devaguptapu2020} predicts the stopping task of an agent during sequential decision-making using Discounted Satisficing (DS) heuristic where the agent's threshold discounts over time. However, the model was found to be inaccurate when people exhibit DS models with varying, yet dependent parameters across days. 

The main contributions of this paper are three-fold. Firstly, this paper proposes a novel decision heuristic called \textbf{\emph{dynamic discounted satisficing}} (DDS) which captures the dynamics of satisficing threshold. 
Secondly, this paper proposes novel \textbf{\emph{neural network architectures with random activations}} that are designed to mimic the DDS heuristic. Thirdly, a novel learning algorithm called \textbf{\emph{sampling-based backpropagation through time (S-BPTT)}} is also developed to train the proposed stochastic neural networks to accurately predict driver's cognitive atrophy via learning the DDS model parameters in a data-driven manner.
The proposed approach is validated on simulation experiments and on Chicago taxi dataset \cite{chicago_2023}. 


The remainder of the paper is organized as follows. Section \ref{sec: DDS model} models driver's cognitive atrophy using dynamic discounted satisficing, and justifies the need for such a model along with an detailed illustrative example.  In Section \ref{sec: Neural Network}, the architecture of the proposed model and a novel Sampling-based Back Propagation Through Time (SBPTT) training algorithm are deliberated upon. Following this, Section \ref{sec: Experiments} elucidates the assumptions regarding model parameters for simulation experiments, as well as the preprocessing steps undertaken for the real-world dataset. The discussion on validation results is contained within Section \ref{sec: Results}. Finally, Section \ref{sec: Conclusion} outlines our future research.

\section{Modeling Driver's Cognitive Atrophy \label{sec: DDS model}}

\subsection{Dynamic Discounted Satisficing}
Consider a ridesharing platform, where a driver serves a total of $T_d$ rides on $d^{th}$ day. Let $u_{d,k}$ denote the utility obtained by the driver upon completing the $k^{th}$ ride on $d^{th}$ day. The total accumulated utility of the driver after completing $T_d$ rides on day $d$ is defined as
\begin{equation}
U_{d,T_d} = \displaystyle \sum_{k=1}^{t} u_{d,k}
\label{Eqn: Accumulated Utility after t rides}
\end{equation}
Henceforth, for simplicity, ignore the subscript $T_d$ in $U_{d,T_d}$ and denote the total utility accumulated by the driver on day $d$ as $U_d$.
The driver's bounded rationality is modeled using \emph{dynamic discounted satisficing} (DDS) which models two different types of dynamics within the satisficing threshold: (i) the attrition of threshold within a given day due to increasing weariness over time, and (ii) the evolution of the initial target and fatigue rate across days. Then, DDS can be formally defined as:

\begin{definition}

A driver is said to exhibit \emph{dynamic discounted satisficing heuristic}, if there exists four real numbers $a_1, a_2, b_1, b_2 \in \mathbb{R}$, one positive real number $\lambda \in \mathbb{R}_+$, one bounded real number $\beta \in (0,1]$, and two arrays of random numbers $\epsilon_d \sim \mathcal{N}(0,1)$ and $\eta_d \sim \mathcal{N}(0,1)$ for $d \in \mathbb{N}$, such that his/her final ride count $t^*$ is given by
\begin{equation}
t^* = {\text{minimize}} \ \left\{ t \in \mathcal{T}_d \ | \ U_{d,t} = \displaystyle \sum_{k=1}^{t} u_{d,k} \ge \beta_d^{t -1}\cdot \lambda_d \right\}
\label{Eqn: DDS - decision}
\end{equation}
where the dynamics of initial target $\lambda_d$ and the discounting factor $\beta_d$ are respectively given by
\begin{equation}
\lambda_d = \mathbb{P}_{[0,\infty]} \Big( a_1\cdot \lambda_{d-1} + a_2 \cdot U_{d-1} + \epsilon_d \Big),
\label{Eqn: DDS - Lambda dynamics}
\end{equation}
and
\begin{equation}
\beta_d = \mathbb{P}_{[0,1]} \Big( b_1\cdot \beta_{d-1} + b_2 \cdot e^{-T_{d-1}} + \eta_d \Big).
\label{Eqn: DDS - Beta dynamics}
\end{equation} 
where $\mathbb{P}_{\mathcal{S}}(x)$ denotes the projection operator that projects the input argument $x$ onto the set $\mathcal{S}$, i.e.
\begin{equation}
\mathbb{P}_{\mathcal{S}}(x) = 
\begin{cases}
x_L, & \text{if } x \leq x_L \triangleq \inf \mathcal{S},
\\[1ex]
x, & \text{if } x \in \mathcal{S},
\\[1ex]
x_U, & \text{if } x \geq x_L \triangleq \sup \mathcal{S}.
\end{cases}
\end{equation}
\vspace{-3ex}
\label{Defn: DDS}
\end{definition}

\begin{figure}[!t]
\centering
\hspace{-2.5ex}
\includegraphics[width=0.495\textwidth]{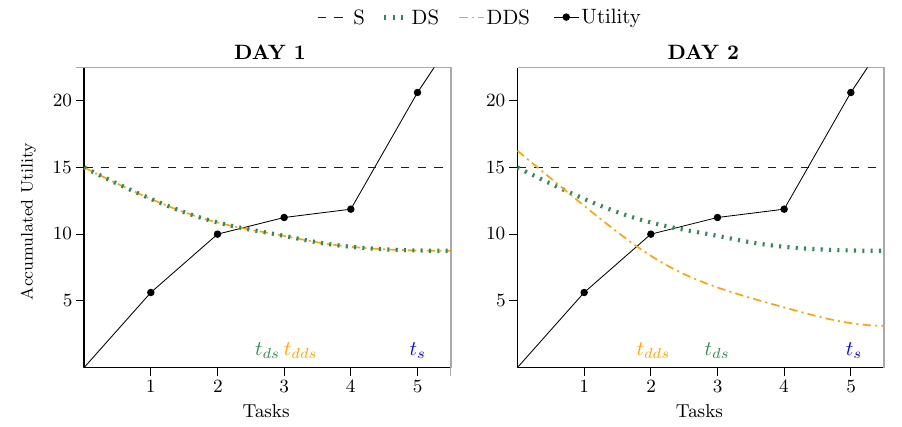}
\caption{Illustrative Example of Three Different Decision-Making Models over 5 Tasks.}
\label{Fig: Illustrative example}
\vspace{-2ex}
\end{figure}

\subsection{Illustrative Example}
Consider an illustrative example of a ride-sharing platform, where a driver has the same set of five taxi rides for two days. Let the corresponding utilities obtained by the driver be given as $U_{d, t} = [6, 4, 2, 0, 9]$, for $d \in \{1, 2\}$ and $t \in \{ 1, \cdots, 5\}$.
Let the driver's DDS model parameters be initialized as $\lambda_0=15$, $\beta_0=0.9$, $a_1=0.8$, $a_2=0.2$, $b_1=0.8$, and $b_2=0.2$. In the case of \emph{Satisficing} ($S$) model, the threshold remains constant at $\lambda_s = 15$ for both days across all tasks. As a result, $S$ yields the same stopping task as $t_s = 5$ for both days, as shown in Figure \ref{Fig: Illustrative example} (Left). On the contrary, assume that the \emph{Discounted Satisficing} ($DS$) model has the threshold and the discounting factor defined as $\lambda_{ds} = 15$ and $\beta_{ds} = 0.9$ respectively. Due to the static nature of $DS$ across days, the stopping task $t_{ds} = 3$ remains the same for both days, just as observed in the case of $S$ model. However, in the proposed DDS model, the threshold and discounting factor varies dynamically across days based on Equations \eqref{Eqn: DDS - Lambda dynamics} and \eqref{Eqn: DDS - Beta dynamics}. Therefore, the threshold and discounting factor on Day-1 are $\lambda_{dds1} = 15$ and $\beta_{dds1} = 0.9$ respectively, which results in a stopping task of $t_{dds} = 3$, while the threshold and discounting factor on Day-2 are $\lambda_{dds2} = 16.2$ and $\beta_{dds2}=0.73$ respectively, thus resulting in a stopping task of $t_{dds} = 2$, as shown in Figure \ref{Fig: Illustrative example}. In other words, since the taxi driver earned more than threshold on Day-1, he/she will more motivated on Day-2. However, with the increase in driver's fatigue, the threshold will decrease steeply over Day-2. As a result, the taxi driver tends to stop serving rides earlier than Day-1.

Note that the $S$ and $DS$ models yield the same stopping times for both days, when the utility of the driver remains constant. However, the proposed DDS model produces a different stopping times for each day, due to the changing threshold and discounting factor. The dynamic nature of the DDS model is consistent with the behavior of ridesharing drivers, especially as their wealth and attrition simultaneously increases across days in most cases.

\subsection{Need for DDS Model}
Satisficing (S) and Discounted Satisficing (DS) are considered as different decision-making models that incorporate the concept of satisficing, i.e., a minimum acceptable level of utility. However, these decision-making models differ in their assumptions on how the threshold is applied and updated over time. The concept of \emph{satisficing} was first introduced by Simon in \cite{simon1958, simon1957}, where it is defined as an agent’s decision to stop choosing from the alternatives when the total accumulated utility goes beyond a specific threshold.
However, satisficing assumes that the threshold constructed by an agent remains constant throughout the decision-making process. Specifically, the objective of the agent is to achieve an accumulated utility that is at least as good as their threshold, but are indifferent towards the optimal outcome. Devaguptapu et al. \cite{Devaguptapu2020} predicts the stopping time of an agent based on discounted satisficing heuristic, where the agent's threshold gets discounted with time. In other words, the agent is assumed to experience discontent as well as discounts their threshold with time, making the agent satisfied much earlier than intended.

Therefore, this paper proposes a novel decision heuristic known as \emph{Dynamic Discounted Satisficing} (DDS) which captures the dynamics of the satisficing threshold. Specifically, the proposed approach allows to capture the change in an agent's discontent/motivation as they approach towards the end of the task over the course of a day.
This paper strongly believes that humans do not always make optimal decisions, but rather aim for satisfactory outcomes given limited information and cognitive resources. The DDS model incorporates temporal dynamics by modeling how the driver's initial target and fatigue rate evolve over time. This is important because real-world decision-making often involves adapting to changing circumstances and learning from past experiences.The DDS model is flexible and can be adapted to different contexts. In this paper, we model the driver's bounded rationality using DDS model and predict how many ride requests a driver will accept on a given day. DDS can improve planning, designing and optimizing ridesharing platforms 
so as to promote more favorable driver behavior.

\section{Modeling DDS using Neural Networks \label{sec: Neural Network}}

\begin{figure}[!t]
\centering
\includegraphics[width=0.475\textwidth]{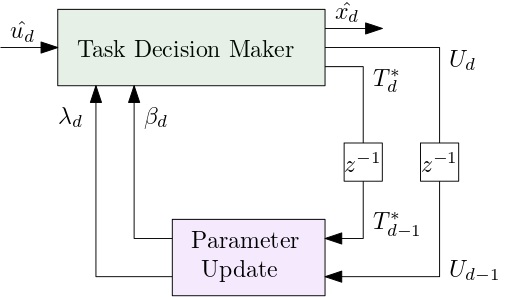}
\caption{Model Architecture}
\vspace{-3ex}
\label{fig:MA}
\end{figure}

\begin{figure}[!t]
\centering
\includegraphics[width=0.495\textwidth]{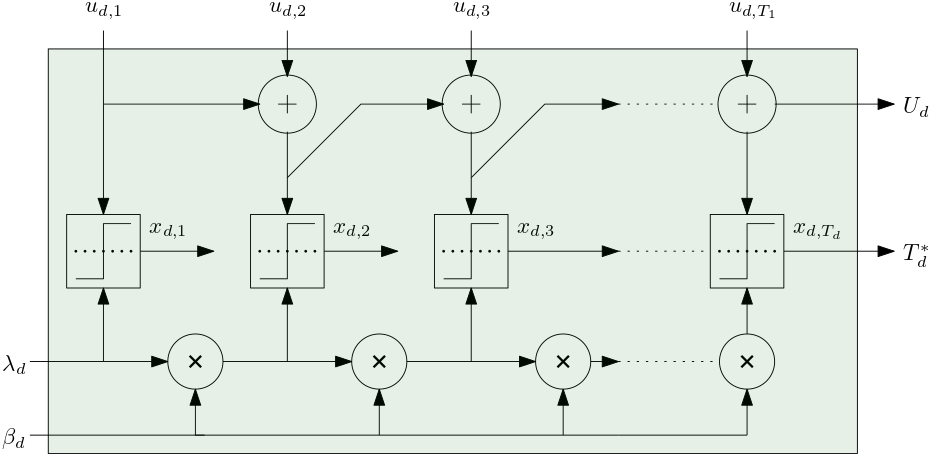}
\caption{Task Decision Maker Architecture}
\label{fig:TDM}
\vspace{-2ex}
\end{figure}

This paper employs the dynamic discounted satisficing model based on the combination of classical statistical modeling techniques and data-driven systems as discussed by Shlezinger et al. in \cite{shlezinger2023model}.
As shown in Figure \ref{fig:MA}, the dynamic discounted satisficing heuristic is modelled as a sequential decision-making strategy employed by a driver, incorporating two vital components: (i) the \emph{task decision maker} and (ii) the \emph{parameter update network}.

\begin{figure}[!t]
\centering
\includegraphics[width=0.425\textwidth]{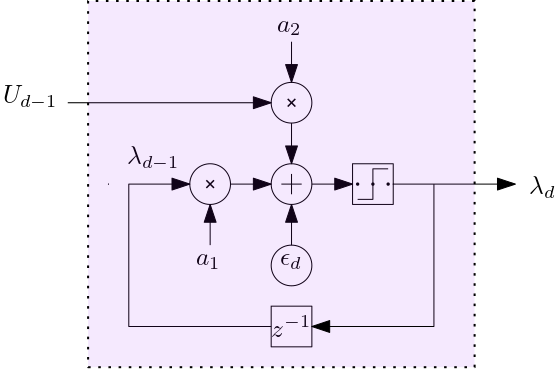}
\caption{Dynamic Threshold Architecture}
\label{fig:PDa}
\vspace{-2ex}
\end{figure}

\begin{figure}[!t]
\centering
\includegraphics[width=0.425\textwidth]{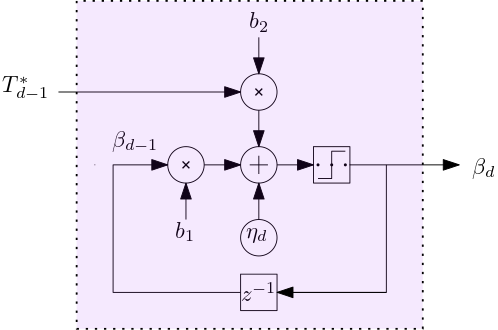}
\caption{Dynamic Threshold Architecture}
\label{fig:PDb}
\vspace{-4ex}
\end{figure}

\begin{algorithm}[!t]
\caption{Sampling-based Back Propagation Through Time}
\label{alg:sbbptt}
\begin{algorithmic}[1]
\Inputs{$\vu*{U}_{D \cross T},  x^*_{D \cross T}, \epsilon \sim \mathcal{N}(0,1), \eta \sim \mathcal{N}(0,1)$}
\vspace{1ex}
\Initialize{\strut Model parameters $w = [a_{1}, a_{2}, b_{1},  b_{2}]$ \\
Learning rate $\alpha = 0.01$ \\
Initial gradient $\displaystyle \frac{\partial \mathcal{L}}{\partial w} = 0$}
\vspace{1ex}
\For{$d=D$ to $1$}
\For{$r=1$ to $R$}
\State Predicted output: $\displaystyle \hat{x}_{d,r} = f(\vu*{U}_{d \cross T}, x^*_{D \cross T},\epsilon_r,\eta_r)$
\State Loss: $\displaystyle L_{d,r} = \mathcal{L}(\hat{x}_{d,r}, x^*_{D \cross T})$
\State Loss gradients: $\displaystyle \frac{\partial \mathcal{L}_{d,r}}{\partial w}$
\EndFor
\State Empirical mean of loss gradients: $\displaystyle \frac{\partial L}{\partial w} = \frac{1}{R} \sum_{r=1}^{R} \frac{\partial \mathcal{L}_{d,r}}{\partial w}$
\State Gradient descent update: $\displaystyle w \leftarrow w - \alpha \frac{\partial \mathcal{L}}{\partial w}$
\EndFor
\end{algorithmic}
\end{algorithm}

\subsection{Architecture Design}
The objective of the Task Decision Maker is to predict the probability that the driver decides to either continue working or stop working for the day. 
Let $x_{d,t}$ denote the probability of the driver choosing to accept the $t^{th}$ ride request on day $d$. As per Definition \ref{Defn: DDS}, the driver continues to accept ride requests as long as the difference between the discounted threshold $\displaystyle \beta_d^{t-1}\cdot\lambda_d$ and the accumulated utility $U_{d,t}$ up to task $t$ on day $d$ is non-negative. In other words, given the driver's utility across $t$ tasks and the dynamically updated parameters, $\lambda_d, \beta_d$, the probability $x_{d,t}$ is defined as follows.
\begin{equation}
x_{d,t} = \sigma \left[\beta_d^{t -1}\cdot \lambda_d - \sum_{k=1}^{t} u_{d,k} \right]
\label{Eqn: Driver Decision Probability}
\end{equation}
If $x_{d,t} \geq 0.5$, the driver will continue to work, otherwise, the driver will stop working on the tasks for the day.

On the other hand, the Parameter Update Network updates $\lambda_d$ and $\beta_d$ as shown in Equation \eqref{Eqn: DDS - Beta dynamics} and \eqref{Eqn: DDS - Lambda dynamics}. The initial threshold $\lambda_d$, is a positive real number, which depends on the previous day $\lambda_{d-1}$, the total accumulated utility of the driver up to the previous day $U_{d-1}$, and $\epsilon_d$ with a standard normal distribution $\epsilon_d \sim \mathcal{N}(0,1)$, as shown in Figure \ref{fig:PDa}. The neural network model used is a noisy perceptron, which introduces randomness into the model. The dynamics of the parameter update network are updated, where $a_1$ and $a_2$ are the model parameters. The model uses the rectified linear unit (ReLU) activation function, which maps input values to the set [0, $\infty$). The discounting behavior of the driver over time is determined by the discounting factor $\beta_d$, where $\beta_d \in (0,1]$. The value of $\beta_d$ on day $d$, as shown in Figure \ref{fig:PDb}, depends on the previous day's discounting factor, $\beta_{d-1}$, the stopping task of the driver on the previous day, $T_{d-1}$, and $\eta_d$ with a standard normal distribution, $\eta_d \sim \mathcal{N}(0,1)$. The neural network model used is a noisy perceptron, where $\eta_d$ introduces randomness into the model. The dynamics of the parameter update network are updated, where $b_1$ and $b_2$ are the model parameters. The model uses the sigmoid activation function, which maps input values to the set [0,1).

\subsection{Training Algorithm and Performance Metrics}
In order to effectively train our neural networks on sequential data, a novel approach called Sampling-based Backpropagation Through Time (SBPTT), inspired by the conventional Backpropagation Through Time (BPTT) algorithm \cite{werbos1990backpropagation}, is proposed. The SBPTT algorithm (Algorithm \ref{alg:sbbptt}) takes $\vu*{U}_{D \cross T}$ as an input sequence and generates a prediction sequence $\hat{x}_{d,r}$. As shown in Algorithm \ref{alg:sbbptt}, the forward propagation is performed to compute the predicted output $\hat{x}_{d,r}$ for each random sample $r$ by passing the input sequence $\vu*{U}_{d \cross T}$ through the network, incorporating random variables $\epsilon_r$ and $\eta_r$ sampled from a normal distribution. Subsequently, the loss $L_{d,r}$ between the predicted output $\hat{x}_{d,r}$ and the true output $x^*_{D \cross T}$ is calculated. Then, the gradients of the loss with respect to the network parameters are computed for each random sample, and are averaged across all the random samples. Finally, the model parameters $w$ are updated using simple gradient descent with a learning rate $\alpha$, and this process is repeated for all time steps $d$ until the model converges to a satisfactory level of accuracy on the training data. Notably, by incorporating randomness through $\epsilon_r$ and $\eta_r$, we account for the randomness in human behavior, potentially enhancing the overall performance of the model.

\section{Experiment Design \label{sec: Experiments}}
One of the main challenges with validating the proposed method on real datasets is our inability to observe cognitive parameters such as the discounting factor $\beta$ and the threshold $\lambda$. Therefore, the proposed method is validated on both simulated experiments as well as real-world datasets. The main goal of the simulation experiments is to validate if the proposed neural network tracks the dynamics within the dynamic discounted satisficing model. Subsequently, the same method is also validated on a real dataset when the DDS model parameters are unobservable. Both these experiments are discussed in detail in the remainder of this section.

\subsection{Simulation Experiment}

A sequence of ride-requests received by a single driver with DDS model parameters $\lambda_0 = 70$, $\beta_0 = 0.87$ is simulated over a period of $N = 500$ days in this experiment. The corresponding ride utilities $\vu*{U}_{D \cross T}$ are randomly generated from an exponential distribution with $scale = 10$. The exponential distribution is a suitable choice for modeling the driver's utilities because it is a continuous probability distribution that is commonly used to model the events in a Poisson process. In the context of our example, the exponential distribution captures the randomness and unpredictability of the driver's utility from each ride.
Assuming that the driver accepts the ride request at the start of each day, his/her sequence of ride-acceptance decisions $y^*_{D \cross T}$ are simulated according to the stated DDS model. Note that the rider decision at any given instant is a binary variable, which takes the value $1$ if the driver accepts a ride-request, or takes the value of $0$ otherwise.


\begin{figure*}[!t]
\centering
\begin{subfigure}{0.32\textwidth}
\centering
\includegraphics[width=\textwidth]{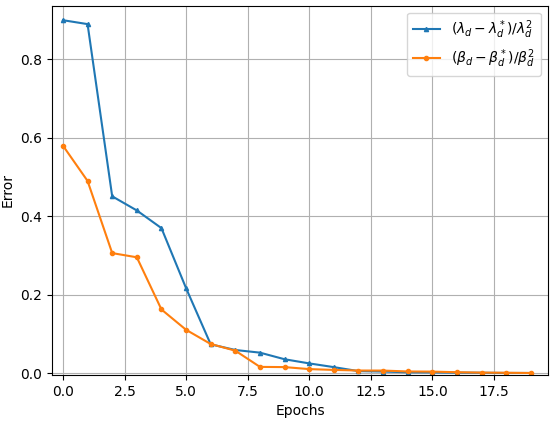}
\caption{$R = 1$}
\label{Fig: Lambda-Beta Error, R=1}
\end{subfigure}
\hfill
\begin{subfigure}{0.32\textwidth}
\centering
\includegraphics[width=\textwidth]{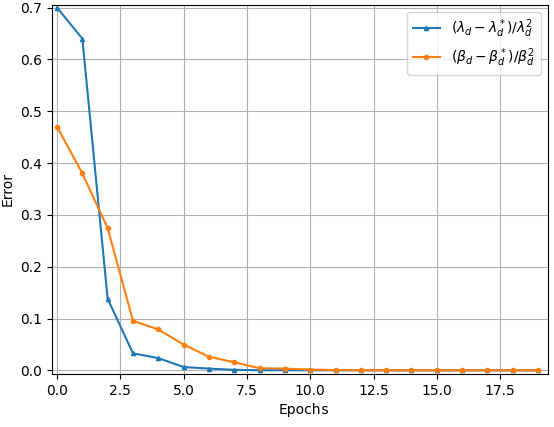}
\caption{$R = 8$}
\label{Fig: Lambda-Beta Error, R=8}
\end{subfigure}
\hfill
\begin{subfigure}{0.32\textwidth}
\centering
\includegraphics[width=\textwidth]{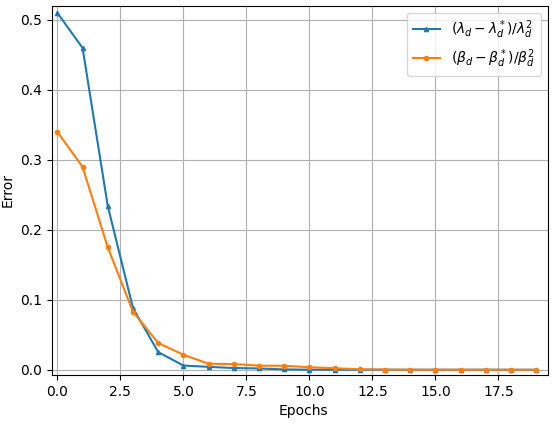}
\caption{$R = 32$}
\label{Fig: Lambda-Beta Error, R=32}
\end{subfigure}
\vspace{-0.5ex}
\caption{Error in Estimating $\lambda_d$ and $\beta_d$ Mode Parameters across 20 Epochs.}
\label{Fig: Lambda-Beta Error}
\vspace{-1ex}
\end{figure*}

\subsection{Chicago Taxi Dataset and Preprocessing}
Our theoretical findings are also validated on a real-world dataset, the Chicago Taxi dataset, that is obtained from the City of Chicago's Open Data Portal \cite{chicago_2023}. This dataset consists information regarding the individual taxi trips served in Chicago including taxi ID, trip start and end timestamps, trip duration, trip distances, trip total fare, and payment type.
For this study, the input attributes are limited to trip start timestamp, trip end timestamp, and trip total fare. Here, the attribute \emph{trip total fare} indicates the amount received by a driver for a specific taxi trip. In other words, the utility of the taxi driver can be defined as the total amount received for all trips served in a day. 
Therefore, the utility of a taxi driver is computed by summing up the trip total fares he/she received for each day. Furthermore, a sample of 10 different taxi drivers is selected randomly.
The resultant dataset is partitioned into two sub-datasets, (i) input dataset, which contains the computed trip total fares for each day, denoted as $\vu*{U}_{D \cross T}$, and (ii) expected output dataset, that contains the same values labeled as expected output, denoted as $y^*_{D \cross T}$. The input dataset is padded with the average of all trip total fares to fill the rest, and the expected output dataset is encoded into binary format, where trip total fare values are replaced with 1 and the remaining with 0. 

The proposed model was trained on the preprocessed dataset with 40-60\% train-test splits. The random variables $\epsilon_d$ and $\eta_d$ are sampled from normal distribution allowing the model to effectively account for the inherent variability and randomness in human behavior, thereby enhancing the overall performance and generalization capabilities. The stochastic nature of the model enables it to effectively capture uncertainties and variations in the data, making it well-suited for predicting the stopping time of taxi drivers in the City of Chicago based on the total amount they received (utility) for each trip. On the other hand, the model parameters $w = [a_1, a_2, b_1, b_2]$ are updated using the gradient descent algorithm with a learning rate of $\alpha = 0.01$. Finally, the Discounted Satisficing (DS) model proposed by Devaguptapu et al. in \cite{Devaguptapu2020} was also trained on the same preprocessed dataset containing $10$ drivers. The performance of our proposed model SBPTT is compared with the DS model in terms of binary cross-entropy loss (BCE) as well as accuracy.

\begin{figure*}[!t]
\centering
\begin{subfigure}{0.45\textwidth}
\includegraphics[width=\textwidth]{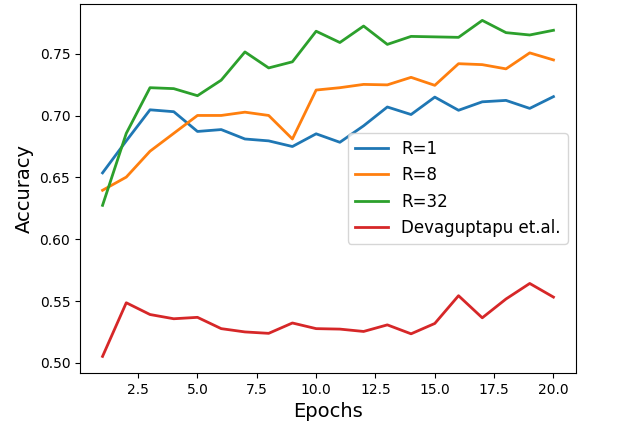}
\caption{Results on Training Data}
\label{fig:SIM_TrainA} 
\end{subfigure}
\hfill
\begin{subfigure}{0.45\textwidth}
\centering
\includegraphics[width=\textwidth]{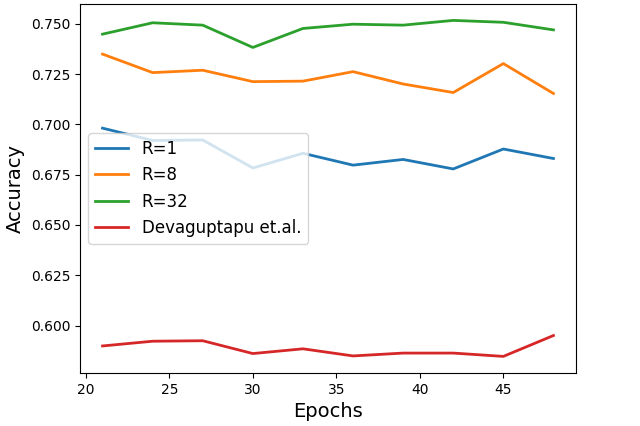}
\caption{Results on Test Data}
\label{fig:SIM_TestA}
\end{subfigure}
\caption{Accuracy of the Proposed Model on Simulated Dataset in Comparison with DS Model.}
\label{Fig: SIM_Accuracy Results}
\vspace{-1ex}
\end{figure*}


\section{Results and Discussion \label{sec: Results}}

\subsection{Simulation Results}
The convergence of the threshold $\lambda_d$ and the discounting factor $\beta_d$ is first validated, given that these model parameters are observable in the simulation setting. Figure \ref{Fig: Lambda-Beta Error} illustrates that the error in $\lambda_d$ and $\beta_d$ converges to zero over 20 epochs for all random sample sizes $R = 1, 8, 32$. More specifically, the proposed model only needs 8 epochs in estimating the model parameters, $\lambda$ and $\beta$, precisely. This can be attributed to the fact that the dynamic thresholds of the driver with higher values of $\beta_d$ generally deteriorate at a much slower rate, thereby revealing about the model parameters in their choices. Moreover, a slight improvement in convergence rate is observed as the number of random samples increase from $R = 1$ to $R = 8, 32$, as shown in Figure \ref{Fig: Lambda-Beta Error, R=32}.

\begin{figure*}[!t]
\centering
\begin{subfigure}{0.45\textwidth}
\includegraphics[width=\textwidth]{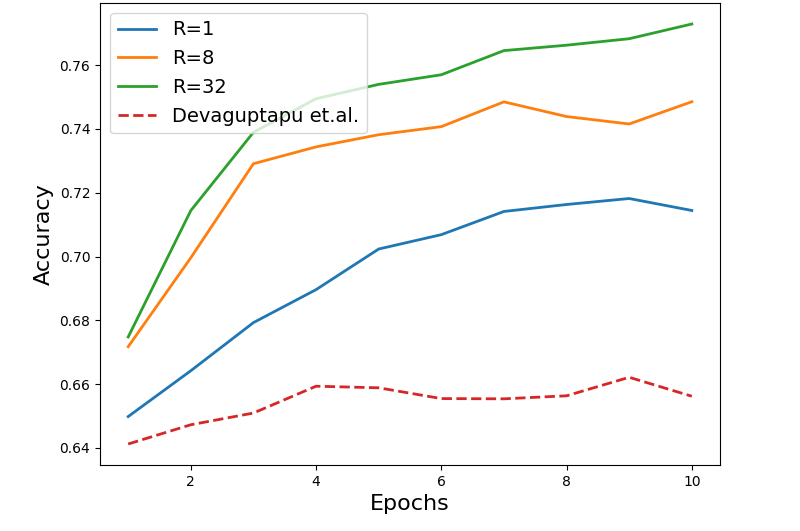}
\caption{Results on Training Data}
\label{Fig: Train Chicago} 
\end{subfigure}
\hfill
\begin{subfigure}{0.45\textwidth}
\centering
\includegraphics[width=\textwidth]{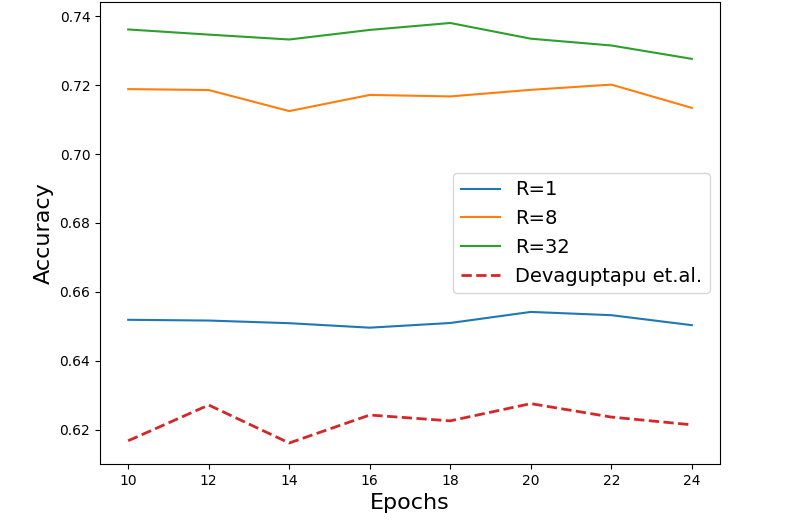}
\caption{Results on Test Data}
\label{Fig: Test Chicago}
\end{subfigure}
\caption{Accuracy of the Proposed Model on Chicago Taxi Dataset in Comparison with DS Model.}
\label{Fig: Chicago Accuracy Results}
\vspace{-2ex}
\end{figure*}


On the hand, Figure \ref{Fig: SIM_Accuracy Results} demonstrates the accuracy of the proposed model in comparison with the DS model. It can be observed that the DDS algorithm outperforms the DS model by almost 25\% in terms of training accuracy, as shown in the Figure \ref{fig:SIM_TrainA}. Moreover, the training accuracy for the DDS algorithm increases with the number of random samples $R = 1, 8, 32$. Similar results are also observed in testing phase as well, where the test accuracy increases with the number of random samples, as shown in Figure \ref{fig:SIM_TestA}. In addition, Figure \ref{Fig: Train Loss - Simulation} depicts the BCE loss incurred while training the simulated data using the proposed algorithm DDS. Specifically, DDS demonstrated less loss when compared to the DS model, regardless of the number of random samples considered. However, on the contrary, the loss incurred with $R = 1$ random sample is less than the loss with $R = 8, 32$ random samples both in training as well as testing phase. 


\subsection{Results on Chicago Taxi Dataset}

\begin{figure*}[!t]
\centering
\begin{subfigure}{0.45\textwidth}
\includegraphics[width=\textwidth]{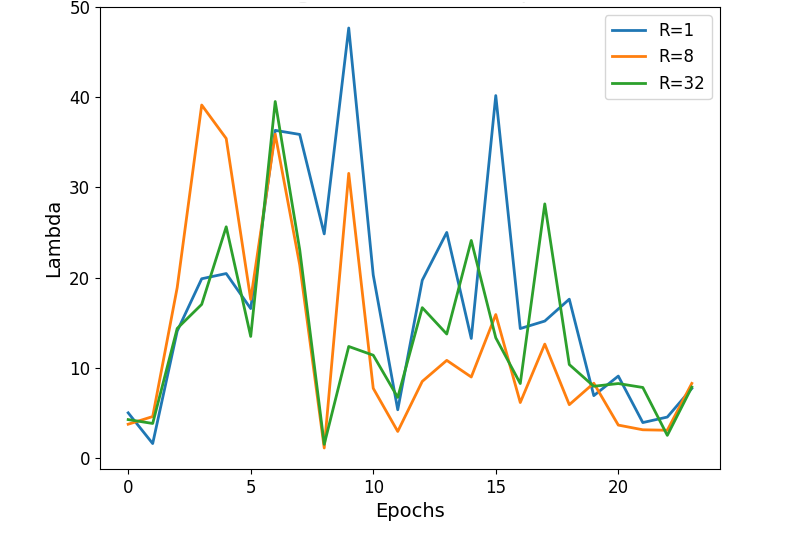}
\caption{Average $\lambda$ for 10 drivers across Epochs}
\label{fig:L}
\end{subfigure}%
\hfill
\begin{subfigure}{0.45\textwidth}
\includegraphics[width=\textwidth]{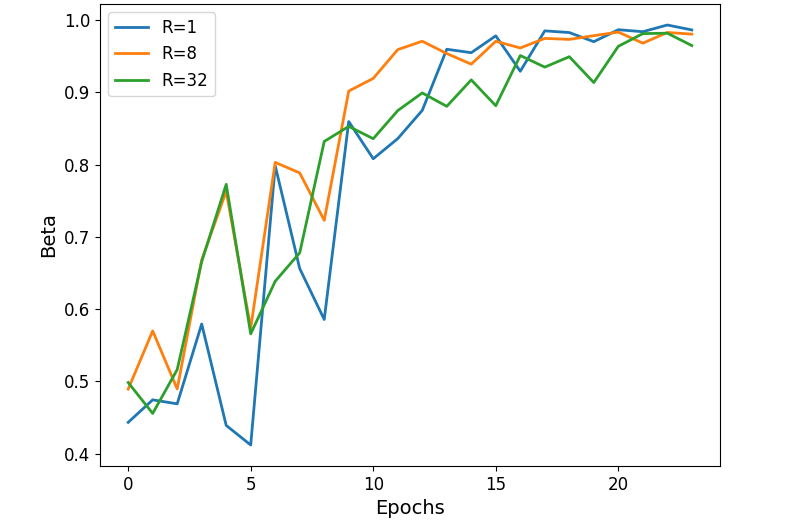}
\caption{Average $\beta$ for 10 drivers across Epochs}
\label{fig:B}
\end{subfigure}
\caption{Average $\lambda$ and $\beta$ across epochs}
\label{fig:LB}
\vspace{-2ex}
\end{figure*}

Figure \ref{Fig: Chicago Accuracy Results} shows the accuracy of the proposed model on Chicago Taxi dataset for different number of random samples $R = 1, 8, 32$. The proposed DDS model attained 78\% accuracy with $R = 32$ samples over 10 epochs, while the accuracy of the DS model \cite{Devaguptapu2020} is around 65\% for the same training dataset, as shown in Figure \ref{Fig: Train Chicago}. Moreover, the accuracy of the DDS model increases with the number of random samples. Similar results can also be observed in testing phase as depicted in Figure \ref{Fig: Test Chicago}. 
In addition, the Figures \ref{fig:L} and \ref{fig:B} demonstrates the average $\lambda$ and $\beta$ values of 10 drivers for different random sample sizes $R$, where it can be implied that as $R$ increases the average $\lambda$ and $\beta$ values become more stable. This stability in the parameters suggests that the model becomes more robust and less sensitive to the specific training data it is exposed to. In summary, the results demonstrate that increasing the value of $R$ has a positive impact on both the loss and accuracy of the model. The trend of decreasing loss and increasing accuracy with higher values of $R$ indicates that the model's performance improves as it learns from more diverse training data. These findings suggest that increasing the number of samples, as represented by the parameter $R$, can lead to better model performance and improved accuracy in predicting the target output. Furthermore, we show that Dynamic Discounted Satisficing model has higher accuracy in predicting the stopping decision of a driver when compared with Discounted Satisficing model.

\section{Conclusion and Future Work \label{sec: Conclusion}}

In this study, we developed a custom neural network model to predict the stopping time of taxi drivers in the City of Chicago based on the total amount paid for each trip they make. Our model achieved an accuracy of $35\%$ and outperformed a baseline model \cite{Devaguptapu2020} that predicted a constant value as the stopping time. In the future, we can improve the model's performance by incorporating additional features such as weather data, traffic congestion data, and the taxi driver's demographics. 
Additionally, we can evaluate the model's performance on a more datasets to further validate its effectiveness in predicting the stopping time of taxi drivers.

\bibliographystyle{ieeetr}
\bibliography{DDS_main}

\end{document}